\def\year{2020}\relax
\documentclass[letterpaper]{article} % DO NOT CHANGE THIS
\usepackage{aaai20}  % DO NOT CHANGE THIS
\usepackage{times}  % DO NOT CHANGE THIS
\usepackage{helvet} % DO NOT CHANGE THIS
\usepackage{courier}  % DO NOT CHANGE THIS
\usepackage[hyphens]{url}  % DO NOT CHANGE THIS
\usepackage{graphicx} % DO NOT CHANGE THIS
\usepackage{graphicx}  % DO NOT CHANGE THIS
\usepackage{algorithm}
\usepackage{algorithmicx}
\usepackage{multirow}
\usepackage{amsmath}
\usepackage{amssymb}
\usepackage{amsfonts}
\usepackage{threeparttable}
\newtheorem{theorem}{Theorem}

\title{
SynSig2Vec: Learning Representations from Synthetic Dynamic Signatures for Real-world Verification
}
\author{Songxuan Lai, Lianwen Jin, Luojun Lin, Yecheng Zhu, Huiyun Mao\\
School of Electronic and Information Engineering, South China University of Technology\\
eesxlai@foxmail.com, eelwjin@scut.edu.cn
}

\begin{document}

\maketitle

\begin{abstract}
An open research problem in automatic signature verification is the skilled forgery attacks. However, the skilled forgeries are very difficult to acquire for representation learning. To tackle this issue, this paper proposes to learn dynamic signature representations through ranking synthesized signatures. First, a neuromotor inspired signature synthesis method is proposed to synthesize signatures with different distortion levels for any template signature. Then, given the templates, we construct a lightweight one-dimensional convolutional network to learn to rank the synthesized samples, and directly optimize the average precision of the ranking to exploit relative and fine-grained signature similarities. Finally, after training, fixed-length representations can be extracted from dynamic signatures of variable lengths for verification. One highlight of our method is that it requires neither skilled nor random forgeries for training, yet it surpasses the state-of-the-art by a large margin on two public benchmarks.

\end{abstract}

\section{Introduction}
\noindent Handwritten signatures are the most socially and legally accepted means of personal authentication. They are generally regarded as a formal and legal means to verify a person’s identity in administrative, commercial and financial applications, for example, when signing credit card receipts. Over the last forty years, research interest in automatic signature verification (ASV) has grown steadily, and a number of comprehensive survey papers have summarized the state-of-the-art results in the field till 2018 \cite{plamondon1989automatic,plamondon2000online,impedovo2008automatic,diaz2019perspective}. Nowadays, building an ASV system to separate genuine signatures and random forgeries (produced by a forger who has no knowledge about the authentic author's name or signature) can be considered a solved task, while to separate genuine signatures and skilled forgeries (produced by a forger after unrestricted practice) still remains an open research problem.

Among the literatures over years, great research effort has been devoted to obtaining good representations for signatures by developing new features and feature selection techniques. With the popularity of deep learning in recent years, several researches have also applied deep learning models to learn representations for dynamic signature sequences or static signature images, and have achieved certain improvements in reducing the verification error against skilled forgeries. However, these methods have several limitations. First of all, they require skilled forgeries as training samples to achieve good performance. One should know that, as a biometric trait and a special kind of private data, handwritten signatures are non-trivial to collect; skilled forgeries, which require the forgers to practise once and again, are even more difficult to acquire. Therefore, these methods can hardly perform equally well when there are only genuine signatures for training. Second, these methods generally lack an appropriate data augmentation method, which is fundamental in training deep learning models. The main reason lies in the following questions: Which type of data augmentation can essentially capture the variance of the underlying signing process? To what extent can an augmented genuine signature maintain its ``genuineness"? Third, existing loss functions do not consider fine-grained signature similarities and tend to overfit. For example, in Siamese networks, positive and negative signature pairs are always labelled with 1s and 0s, respectively, regardless of the actual visual similarities. 

\begin{figure}
  \centering
  \includegraphics[width=.9\columnwidth]{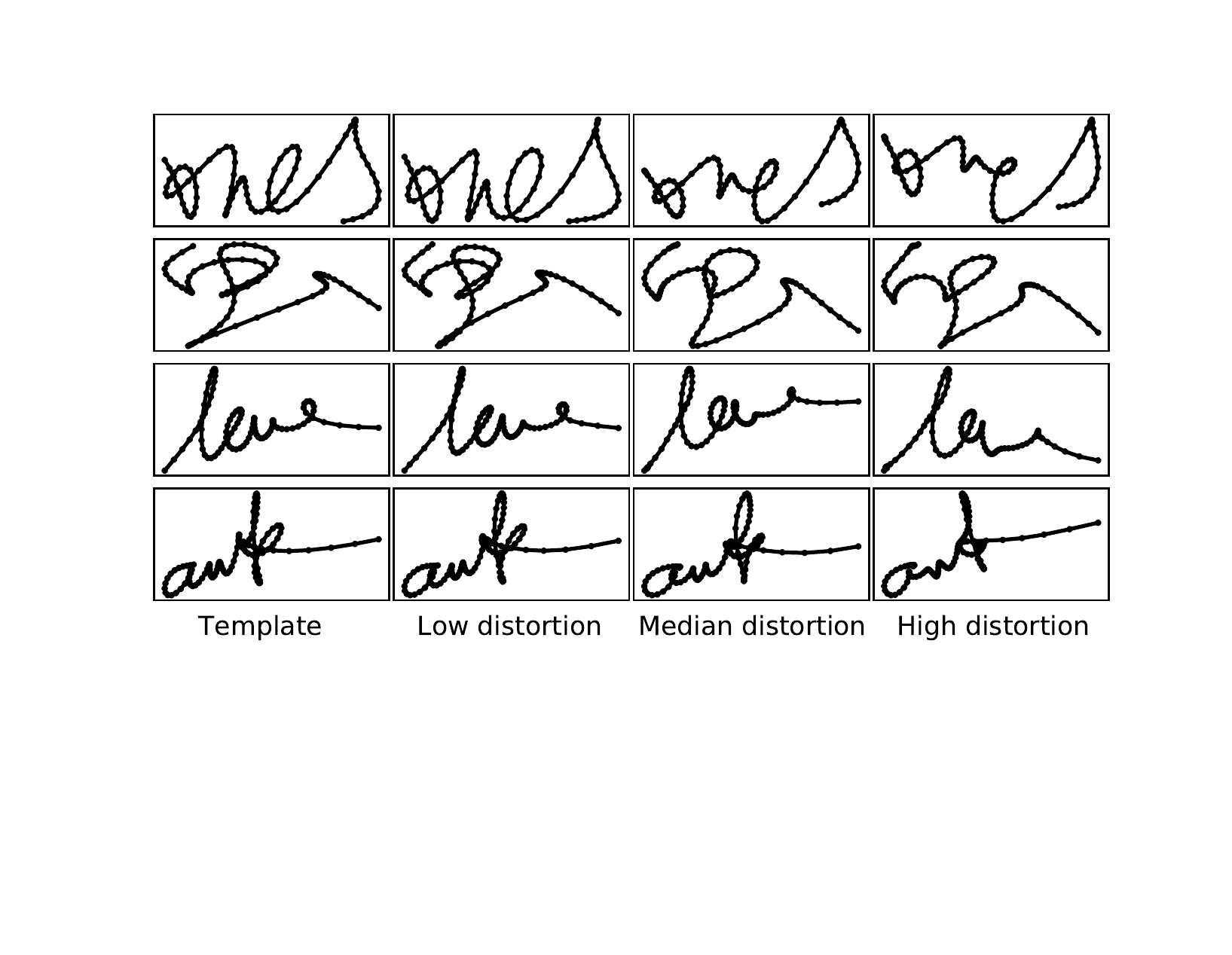}
  \caption{Based on one template signature, the $\Sigma\Lambda$ model is used to synthesize signatures with various distortion levels.}\label{distortionSamples}
\end{figure}

In this paper, we focus on dynamic signature verification and propose a novel ASV system without the above-mentioned limitations. A basis of our method is the kinematic theory of rapid human movements and its Sigma Lognormal ($\Sigma\Lambda$) model \cite{plamondon1995kinematic}. The $\Sigma\Lambda$ model hypothesizes that the velocity of a neuromuscular system can be modelled by a vector summation of a number of lognormal functions, each of which is described by six parameters. Rooted in this model, we extract the underlying neuromuscular parameters of genuine signatures and synthesize new signatures by introducing perturbations to the parameters. The level of parameter perturbation controls the level of signature distortion; based on one genuine signature, we can synthesize signatures with various distortion levels, as shown in Fig. \ref{distortionSamples}. Thereafter, many representation learning methodologies can be considered, such as metric learning. In this study, we propose to learn dynamic signature representations through optimizing the average precision (AP) of signature similarity ranking based on the direct loss minimization framework proposed by \cite{song2016training}. This learning strategy has two benefits. First, as a list-wise ranking method, AP optimization can preserve and exploit fine-grained signature similarities in the ranking list. Second, it is expected to improve the performance since AP is closely related to verification accuracy. Signature similarities are computed as cosine similarities of representations extracted from one-dimensional convolutional neural network (CNN).

The main contributions of this paper are three-fold. First, the application of $\Sigma\Lambda$ model to signature synthesis not only eliminates the need for skilled forgeries, but also serves as a data augmentation technique. Second, we introduce AP optimization and demonstrate its effectiveness for dynamic signature representation learning. Third, we design a simple yet state-of-the-art CNN structure to extract fixed-length representations from dynamic signatures of variable lengths.

\section{Related Work}
Application of deep learning to dynamic signature verification has not been explored too much due to the difficulty of collecting large datasets. Existing studies in the field can be roughly divided into two categories. The first category learns local representations \cite{lai2018recurrent,wu2019prewarping}. It maintains the temporal information of the input and applies dynamic time warping (DTW) to the learned feature sequence; to this end, specific techniques may be needed during training, such as a modified gated recurrent unit \cite{lai2018recurrent} and signature prewarping \cite{wu2019prewarping}. The second category learns fixed-length global representations \cite{tolosana2018exploring,ahrabian2018usage,park2019robust}. For example, \cite{park2019robust} used CNN and time interval embedding to extract features from dynamic signature strokes, and then utilized recurrent neural networks to aggregate over the strokes. As mentioned in the introduction, the above methods require training with skilled forgeries to enhance the performance when verifying this type of samples, which may be impractical in many situations. Our method only requires genuine signatures because of the introduced synthesis method, and learns global representations using a lightweight CNN.

Two relevant studies to ours also use $\Sigma\Lambda$-based synthetic signatures to train dynamic ASV systems. \cite{diaz2016dynamic} synthesized auxiliary template signatures to enhance several non-deep learning systems, including DTW-based, HMM-based and Manhattan-based systems. Whether synthetic signatures are effective in deep learning was not validated in their study. \cite{ahrabian2018usage} trained and tested on fully synthetic signatures using recurrent autoencoder and the Siamese network. Whether synthetic data helps to verify real world signatures was not investigated. Also, different to these two studies, our method learns to rank the signatures synthesized with different distortion levels, which is a novel idea in the field of ASV.

\section{Methodology}
\begin{table*}[tb]
\caption{Configurations of the random variables that decide the signature distortion levels.}
\label{distortionRange}
\centering
\resizebox{.9\textwidth}{!}{
\begin{threeparttable}
\begin{tabular}{cccc}
  \hline
  \multirow{2}{*}{Parameters}&\multirow{2}{*}{Admissible Ranges}&\multicolumn{2}{c}{Distortion Levels}\\
  \cline{3-4}
  &&$\mathcal{G}_1$&$\mathcal{G}_2$\\
  \hline
  $D_i$&$l_{D}$=-0.1000, $h_{D}$=0.1000.\tnote{*}&$R_D\sim U[0.25l_D,0.25h_D]$&$R_D\sim U[0.65l_D,0.40l_D]\cup[0.40h_D,0.65h_D]$\\

  $t_{0_i}$&$l_{t_0}$=-0.0825, $h_{t_0}$=0.0850.&$R_{t_0}\sim U[0.25l_{t_0}, 0.25h_{t_0}]$&$R_{t_0}\sim U[0.65l_{t_0}, 0.40l_{t_0}]\cup[0.40h_{t_0}, 0.65h_{t_0}]$\\

  $\mu_i$&$l_\mu$=-0.3950, $h_\mu$=0.3775.&$R_{\mu}\sim U[0.17l_{\mu}, 0.17h_{\mu}]$&$R_{\mu}\sim U[0.43l_{\mu}, 0.27l_{\mu}]\cup[0.27l_{\mu}, 0.43h_{\mu}]$\\

  $\sigma_i$&$l_\sigma$=-0.3250, $h_\sigma$=0.2875.&$R_{\sigma}\sim U[0.25l_{\sigma}, 0.25h_{\sigma}]$&$R_{\sigma}\sim U[0.65l_{\sigma}, 0.40l_{\sigma}]\cup[0.40l_{\sigma}, 0.65h_{\sigma}]$\\

  $\theta_{s_i}$&Not used&$R_{\theta_s}\sim U[-0.10,0.10]$&$R_{\theta_s}\sim U[-0.20,0.20]$\\

  $\theta_{e_i}$&Not used&$R_{\theta_e}\sim U[-0.10,0.10]$&$R_{\theta_s}\sim U[-0.20,0.20]$\\
  \hline
\end{tabular}
\begin{tablenotes}
  \footnotesize
  \item[*] \cite{bhattacharya2017sigma} did not work on $D$'s admissible ranges. We decide this range based on our visual tests.
\end{tablenotes}
\end{threeparttable}
}
\end{table*}
\subsection{$\Sigma\Lambda$-based Signature Synthesis}
\begin{figure}
  \centering
  \includegraphics[width=0.95\columnwidth]{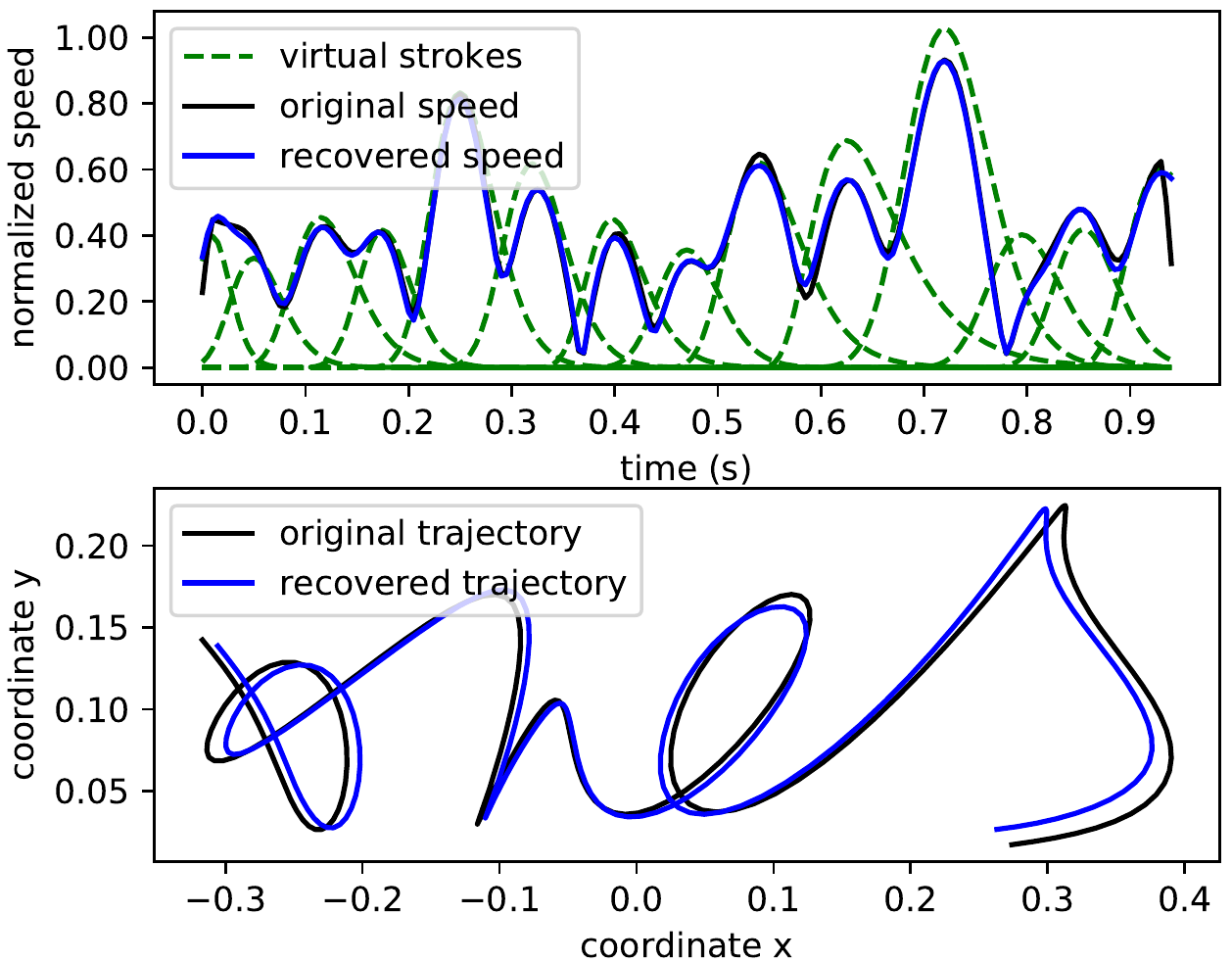} 
  \caption{The velocity profile of a human handwriting component consists of lognormal impulse responses, called strokes. Based on the strokes's parameters, the velocity and trajectory of a component can be recovered.}\label{figLognormal}
\end{figure}

The kinematic theory of rapid human movements, from which the $\Sigma\Lambda$ model was developed, suggests that human handwriting consists in controlling the pen-tip velocity with overlapping lognormal impulse responses, called strokes, as illustrated in Fig. \ref{figLognormal}. The magnitude and direction of the velocity profile $\vec{v}_i(t)$ of the $i$-th stroke is described as:
\begin{equation}\label{speed}
  |\vec{v}_i(t)| = \frac{D_i}{\sqrt{2\pi}\sigma_i(t-t_{0_i})}\exp(\frac{(\ln(t-t_{0_i})-\mu_i)^2}{-2\sigma_i^2}),
\end{equation}
\begin{equation}\label{angle}
  \phi_i(t) = \theta_{s_i}+\frac{\theta_{e_i}-\theta_{s_i}}{D_i}\int^t_0|\vec{v}_i(\tau)|d\tau,
\end{equation}
where $D_i$ is the amplitude of the stroke, $t_{0_i}$ the time occurrence, $\mu_i$ the log time delay, $\sigma_i$ the log response time, $\theta_{s_i}$ and $\theta_{e_i}$ respectively the starting angle and ending angle of the stroke. And the velocity of the complete handwriting movement is considered as the vector summation of the individual stroke velocities:
\begin{equation}\label{SL}
  \vec{v}(t) = \sum_{i=1}^N\vec{v}_i(t),
\end{equation}
$N$ being the number of strokes. In short, each stroke is defined by six parameters: $P_i=[D_i, t_{0_i}, \mu_i, \sigma_i, \theta_{s_i}, \theta_{e_i}]^T$, and a complete handwriting component is defined by $P=[P_1,P_2,...,P_N]^T$.\footnote{The $\Sigma\Lambda$ parameters are extracted based on our implementation of \cite{o2009development}.} In this study, one complete ``component" refers to the trajectory of a pen-down movement, and each component, i.e. pen-down, in the signature is analyzed individually. Although the entire signature can be viewed as a single component by considering the pen-up movements, this practice is less preferred due to two reasons. First, it complicates the parameter extraction process. Second, many current devices and datasets do not record pen-ups.

Based on the extracted parameters in $P$, the component can be reconstructed as follows:
\begin{equation}\label{reconX}
  x(t) = \int_0^t\sum_{i=1}^n|\vec{v}_i(\tau)|cos(\phi_i(\tau))d\tau+\varepsilon_x(t),
\end{equation}
\begin{equation}\label{reconY}
  y(t) = \int_0^t\sum_{i=1}^n|\vec{v}_i(\tau)|sin(\phi_i(\tau))d\tau+\varepsilon_y(t),
\end{equation}
where $\varepsilon_x(t)$ and $\varepsilon_y(t)$ are the residual trajectories that are considered (by the parameter extraction algorithm) to contain no valid strokes. By introducing perturbations to the parameters in $P$, new synthetic components can be generated. A new signature can thus be generated by synthesizing every component in the template signature.

$D_i$, $t_{0_i}$, $\mu_i$ and $\sigma_i$ are distorted as follows:
\begin{equation}\label{D}
  \hat{D}_i=D_i(1+R_D),
\end{equation}
\begin{equation}\label{T}
  \hat{t}_{0_i}=t_{0_i}(1+R_{t_0}),
\end{equation}
\begin{equation}\label{Mu}
  \hat{\mu}_i=\mu_i(1+R_\mu),
\end{equation}
\begin{equation}\label{Sigma}
  \hat{\sigma}_i=\sigma_i(1+R_\sigma),
\end{equation}
while $\theta_{s_i}$, $\theta_{e_i}$ are distorted as follows:
\begin{equation}\label{thetaS}
  \hat{\theta}_{s_i}=\theta_{s_i}+R_{\theta_s},
\end{equation}
\begin{equation}\label{thetaE}
  \hat{\theta}_{e_i}=\theta_{s_i}+R_{\theta_e}.
\end{equation}
$R_D$, $R_{t_0}$, $R_\mu$, $R_\sigma$, $R_{\theta_s}$, $R_{\theta_e}$ are uniform random variables that decide the signature distortion level, and are fixed for all strokes across a component. A previous study \cite{bhattacharya2017sigma} carried out visual Turing test on synthetic characters and worked out the admissible ranges (in percentage terms) of parameter variation: varying parameters out of these ranges will make the character unrecognizable. We borrow the result for $R_{t_0}$, $R_\mu$ and $R_\sigma$; as for $R_D$, $R_{\theta_s}$, $R_{\theta_e}$, they are empirically restricted in a small range. On this basis, for each genuine signature, two groups of signatures, denoted below as $\mathcal{G}_1$ and $\mathcal{G}_2$ respectively, are generated with two different distortion levels as shown in Table I. Signatures in $\mathcal{G}_1$ have lower distortion levels and should rank higher according to the similarity to the template signature, as compared to those in $\mathcal{G}_2$. In this context, one can regard $\mathcal{G}_1$ as augmented genuine signatures, and $\mathcal{G}_2$ as synthetic skilled forgeries. These two distortion levels correspond to the low and median distortion levels as illustrated in Fig. \ref{distortionSamples}, and are determined through a simple coarse grid search, leaving room for future improvements.

To preserve and exploit fine-grained signature similarities, we construct one-dimensional CNNs to learn to rank these synthesized signatures, and optimize the AP of the ranking, as described in the following section.

\subsection{Average Precision Optimization}
Given one genuine signature and its synthetic samples, we compute and rank their similarities and incorporate the AP of the ranking into the loss function for optimization. Because AP is non-differential to the CNN's outputs, we resort to the General Loss Gradient Theorem proposed by \cite{song2016training}, which provides us with the weight update rule for the optimization of AP. Specifically, a neural network can be viewed as a composite scoring function $F(x,y;w)$, which depends on the input $x\in \mathcal{X}$, the output $y\in\mathcal{Y}$, and some parameters $w$. The theorem (cited here for completeness) states that:

\begin{theorem}
When given a finite set $\mathcal{Y}$, a scoring function $F(x,y;w)$, a data distribution, as well as a task-loss $L(y,\hat{y})$, then, under some mild regularity conditions, the direct loss gradient has the following form:
\begin{equation}\label{generalLossGrad}
\begin{aligned}
  &\nabla_w \mathbb{E}[L(y,y_w)]\\
 =&\pm\lim_{\epsilon\to0}\frac{1}{\epsilon}\mathbb{E}[\nabla_wF(x,y_{direct};w)-\nabla_wF(x,y_w;w)],
\end{aligned}
\end{equation}
with
\begin{equation}\label{yw}
  y_w=\mathop{\arg\max}_{\hat{y}\in\mathcal{Y}}F(x,\hat{y};w),
\end{equation}
\begin{equation}\label{ydirect}
  y_{direct}=\mathop{\arg\max}_{\hat{y}\in\mathcal{Y}}F(x,\hat{y};w)\pm\epsilon L(y,\hat{y}).
\end{equation}
\end{theorem}
In Eq. \ref{generalLossGrad}, two directions ($\pm$) of optimization can be used. The positive direction steps away from worse parameters $w$, while the negative direction moves toward better ones.

In the context of this paper, $x$ refers to the concatenation of the two synthetic signature groups for one given genuine signature $x_0$, namely $x=\mathcal{G}_1\cup\mathcal{G}_2=\{x_1,...,x_N\}$, and $N=|\mathcal{G}_1|+|\mathcal{G}_2|$. And $y=\{...,y_{ij},...\}$ is the collection of all pairwise comparisons, where $\forall i,j\in\{1,...,N\}, y_{ij}=1$ if $x_i$ is ranked higher than $x_j$, $y_{ii}=0$, and $y_{ij}=-1$ otherwise. We define the scoring function as follows:
\begin{equation}\label{scoreFucntion}
  F(x,y;w)=\frac{1}{|\mathcal{G}_1||\mathcal{G}_2|}\sum_{\mbox{\tiny$\begin{array}{c}
                                                          x_i\in\mathcal{G}_1\\
                                                          x_j\in\mathcal{G}_2\end{array}$}}y_{ij}(\varphi(x_i;w)-\varphi(x_j;w)),
\end{equation}
where
\begin{equation}\label{cosineSimilarity}
  \varphi(x_i;w)=\frac{f(x_0;w)\cdot f(x_i;w)}{|f(x_0;w)||f(x_i;w)|},
\end{equation}
and $f(\cdot;w)$ is the embedding function parameterized by a CNN with learnable parameters $w$. The scoring function $F(x,y;w)$ is inherited from \cite{yue2007support} and $\varphi(x_i;w)$ measures the cosine similarity of representations from samples $x_0$ and $x_i$. We notice that similar definitions to Eqs. \ref{scoreFucntion} and \ref{cosineSimilarity} have been used for few-shot learning \cite{triantafillou2017few}.

Further, let $p=rank(y)\in\{0,1\}^{|\mathcal{G}_1|+|\mathcal{G}_2|}$ be a vector constructed by sorting the data points according to the ranking defined by $y$, such that $p_j=1$ if the $j$-th data point belongs to $\mathcal{G}_1$ and $p_j=0$ otherwise. Then, given ground truth and predicted configurations $y$ and $\hat{y}$, the AP loss is
\begin{equation}\label{APloss}
  L(y,\hat{y}) \triangleq L_{AP}(p,\hat{p})=1-\frac{1}{|\mathcal{G}_1|}\sum_{j:\hat{p}_j=1}\text{Prec@}j,
\end{equation}
where $\hat{p}=rank(\hat{y})$ and Prec@$j$ is the percentage of samples belonging to $\mathcal{G}_1$ that are ranked above position $j$. To compute the AP loss gradient, we need to infer $y_w$ and $y_{direct}$ in Eqs. \ref{yw} and \ref{ydirect}. For $y_w$, the solution is simple:
\begin{equation}\label{stdInference}
  y_{w_{ij}}=\begin{cases}
            +1, &\text{if } \varphi(x_i;w)>\varphi(x_j;w),\\
        	-1, &\text{otherwise.}
            \end{cases}
\end{equation}
While $y_{direct}$ can be inferred via a dynamic programming algorithm \cite{song2016training}. To prevent overconfidence of the scoring function, we add a regularization term to the AP loss, and obtain the following gradient
\begin{equation}\label{APgrad}
\begin{aligned}
  \nabla_w L_{AP}^{\lambda}=&\pm[\nabla_wF(x,y_{direct};w)-\nabla_wF(x,y_w;w)]\\
                  &+\lambda\nabla_wF(x,y_{direct};w).
\end{aligned}
\end{equation}
Besides the AP loss, we also employ the standard cross-entropy loss $L_{CE}$ for signature classification. Intuitively, these two losses protect the ASV system from skilled forgery and random forgery attacks, respectively. The overall optimization process is given in algorithm 1.

\begin{algorithm}[t]
\caption{Network optimization for learning dynamic signature representations}
\hspace*{0.02in} {Repeat until converged:}
\begin{algorithmic}[1]
\State Draw a genuine signature $x_0$ from a random class (i.e., each individual is a different class) and synthesize $\mathcal{G}_1$ and $\mathcal{G}_2$ based on Eqs. 4-11.
\State Forward pass of network.
\State Infer $y_w$ and $y_{direct}$ in Eqs. \ref{yw} and \ref{ydirect}.
\State Compute AP loss gradient $\nabla_w L_{AP}^{\lambda}$ in Eq. \ref{APgrad}, and cross-entropy loss gradient $\nabla_w L_{CE}$.
\State Update network: $w\leftarrow w-\alpha(\nabla_w L_{AP}^{\lambda}+\nabla_w L_{CE})$
\end{algorithmic}
\end{algorithm}
\subsection{Framework of SynSig2Vec ASV Method}
In this part, we describe our overall framework, including signature preprocessing, CNN structure, and how we construct the verifier based on CNN representations.

\textbf{Signature synthesis and preprocessing.} For signature synthesis, each signature component is resampled at 200 Hz, which is the suggested sampling rate for $\Sigma\Lambda$ parameter extraction \cite{o2009development}. Then, a Butterworth lowpass filter with a cutoff frequency of 10 Hz is applied to the resampled trajectory to enhance the signal. The filtered component is then used for parameter extraction; based on extracted parameters, a synthetic component is generated and resampled at 100 Hz, which is the sampling rate of most existing dynamic signature datasets. As for real handwritten signatures used in this study (all collected at 100 Hz), they are also filtered with the Butterworth lowpass filter to be consistent with synthetic ones.

As we have omitted the pen-up components, we use a straight line to connect the end of a pen-down component and beginning of the next one. These lines can be viewed as virtual pen-ups and have a constant speed equaling the average speed of the pen-downs. Because the essential difference between genuine signatures and the synthetic ones lies in their velocity profiles, we extract three feature sequences as follows:
\begin{equation}\label{vx}
  v_x[n] = \frac{x[n+1]-x[n-1]}{2},
\end{equation}
\begin{equation}\label{vy}
  v_y[n] = \frac{y[n+1]-y[n-1]}{2},
\end{equation}
\begin{equation}\label{v}
  v[n] = \sqrt{v_x[n]^2+v_y[n]^2},
\end{equation}
where $x[n]$ and $y[n]$ are the coordinate sequence. These three feature sequences are normalized to have zero mean and unit variance, and then used as inputs for CNN.

\textbf{Network structure.} A one-dimensional CNN with six convolution layers and scaled exponential linear units (SELUs) \cite{klambauer2017self}, as shown in Table 2, is employed to learn fix-length representations from signature sequences. Batch normalization is not applied, because during training each batch consists of only one genuine signature and its synthesized samples, which are non-i.i.d samples. Nevertheless, SELU provides an alternative normalization effect, and is found to work surprisingly well in our study. Because the signature length may vary after synthesis, we pad all signatures inside a batch with zeros to the maximal length. And a corresponding mask is generated to perform masked average pooling from feature sequences coming out of the sixth convolutional layer.

The receptive field of the sixth convolutional layer is 54. For dynamic signatures sampled at 100 Hz, such a receptive field covers a time interval of 0.5 second and captures several lognormal strokes. We have experimented with deeper networks and larger receptive fields, but found no significant further improvements. We have also explored the residual connections, but found they degraded the performance.

A 256-dimensional feature vector is obtained from the masked average pooling layer, and then goes into two branches. The first branch is a fully connected layer with softmax activation, and the cross-entropy loss is computed on top of it. The second branch is a fully connected layer with 512 neurons, on top of which the AP loss is computed. The AP loss and the cross-entropy loss are used together to optimize the network parameters as described in Algorithm 1. After training, only the second branch is kept. Then, for any given dynamic signature, a 512-dimensional feature vector can be extracted as its representation.

\begin{table}[tb]
\caption{Our CNN architecture. All convolutional layers are followed by SELUs. FC stands for ``fully connected", and k, s, p are kernel, stride and padding sizes, respectively.}
\label{CNN}
\centering
\begin{tabular}{|c|c|c|}
  \hline
  $n$-th layer & \multicolumn{2}{|c|}{Configurations} \\
  \hline
  1 & \multicolumn{2}{|c|}{1D Convolution, 64, k7, s1, p3}\\
  \hline
  2 & \multicolumn{2}{|c|}{1D Max Pooling, k2, s2}  \\
  \hline
  3 & \multicolumn{2}{|c|}{1D Convolution, 64, k3, s1, p1} \\
  \hline
  4 & \multicolumn{2}{|c|}{1D Convolution, 128, k3, s1, p1} \\
  \hline
  5 & \multicolumn{2}{|c|}{1D Max Pooling, k2, s2} \\
  \hline
  6 & \multicolumn{2}{|c|}{1D Convolution, 128, k3, s1, p1} \\
  \hline
  7 & \multicolumn{2}{|c|}{1D Convolution, 256, k3, s1, p1} \\
  \hline
  8 & \multicolumn{2}{|c|}{1D Max Pooling, k2, s2} \\
  \hline
  9 & \multicolumn{2}{|c|}{1D Convolution, 256, k3, s1, p1} \\
  \hline
  10 & \multicolumn{2}{|c|}{Masked 1D Average Pooling} \\
  \hline
  \multirow{2}{*}{11} & FC with softmax & \multirow{2}{*}{FC, 512} \\
     & activation, $M$ classes & \\
  \hline
  & Cross-entropy & AP loss\\
  \hline
\end{tabular}
\end{table}

\textbf{Verifier.} We use a distance-based verifier with the same normalization technique as in \cite{lai2018recurrent}. Specifically, given two signatures $x_i$ and $x_j$, we compute the Euclidean distance of their $l_2$ normalized vectors:
\begin{equation}\label{distance}
  d(x_i,x_j)=\|\frac{f(x_i;w)}{|f(x_i;w)|}-\frac{f(x_j;w)}{|f(x_j;w)|}\|_2,
\end{equation}
where $f(x_i;w)$ is the 512-dimensional feature vector extracted from the CNN. Given $n$ template signatures $\{x^k_1,...,x^k_n\}$ from client $k$, we compute the average pair-wise distances as $\bar{d}_k$ ($\bar{d}_k=1$ if $n=1$). Then, for a test signature $x^{test}$ claimed to be client $k$, we compute the following scores:
\begin{equation}\label{verificationScore}
  s(x^k_i,x^{test})=d(x^k_i,x^{test})/\sqrt{\bar{d}_k}, \forall i\in \{1,...,n\}.
\end{equation}
From these scores, the average score $s_{ave}$ and minimum score $s_{min}$ are computed, leading to a 2-D scatter plot on which we fit a line and make the decision. In practice, we find simple $X+Y=c$ already works well. By varying the threshold $c$, we can obtain the equal error rates (EERs) to assess the system performance. Unless mentioned otherwise, a global threshold for all individuals is used.

\section{Experiments}
\subsection{Datasets and protocols}
Two benchmark dynamic signature datasets were used in this study, namely MCYT-100 \cite{ortega2003mcyt} and SVC-Task2 \cite{yeung2004svc2004}. The MCYT-100 dataset consists of 100 individuals with 25 genuine signatures and 25 skilled forgeries for each individual. The SVC-Task2 dataset contains 40 individuals with 20 genuine and 20 forged signatures per individual.

For MCYT-100, we used a 10-fold cross validation. The $k$-th fold corresponded to the $k$-th ten individuals, and we trained the models in a round-robin fashion on all but one of the folds. Skilled forgeries were not included in the training set. In the testing stage, we considered two scenarios, namely T5 and T1. In scenario T5, five genuine signatures were randomly selected as templates for each individual in the test fold, while the rest 20 genuine and 25 forged signatures were used for testing; EERs were computed and averaged over 50 trials. In scenario T1, each genuine signature was considered as a single template to test against the rest signatures. Finally, for both scenarios, EERs were averaged again over 10 test folds.

For SVC-Task2, similarly we used a 10-fold cross validation and considered scenarios T5 and T1. The training set only included $36\times20=720$ real signatures, therefore both the network and learning algorithm should be data-efficient.
\subsection{Implementation details}
For signature synthesis, to accelerate training, synthetic signatures were first generated offline to create two data pools $\mathcal{P}_1$ and $\mathcal{P}_2$ for each genuine signature. Then, during training, $\mathcal{G}_1$ and $\mathcal{G}_2$ were drawn from $\mathcal{P}_1$ and $\mathcal{P}_2$ respectively. We set $|\mathcal{P}_1|=|\mathcal{P}_2|=20$, $|\mathcal{G}_1|=5$ and $|\mathcal{G}_2|=10$. Therefore, the batch size was $1+|\mathcal{G}_1|+|\mathcal{G}_2|=16$.

For AP optimization, we chose the positive direction and $\lambda$ was set as 5 for MCYT-100 and 10 for SVC-Task2. A larger value of $\lambda$ for SVC-Task2 led to better generalization because of the small dataset scale. The models were optimized using stochastic gradient descent. The learning rate, momentum and weight decay were set as 0.001, 0.9 and 0.001, respectively. We trained for $M\times800$ batches, where $M$ was the number of classes in the training set (90 for MCYT-100 and 36 for SVC-Task2). For the proposed method, the final models were evaluated to report the EERs. For the methods to be compared, the best models were evaluated to see their capacities.
\subsection{Results}
\begin{figure*}[tb]
  \centering
  \includegraphics[width=0.8\textwidth, height=0.35\textwidth]{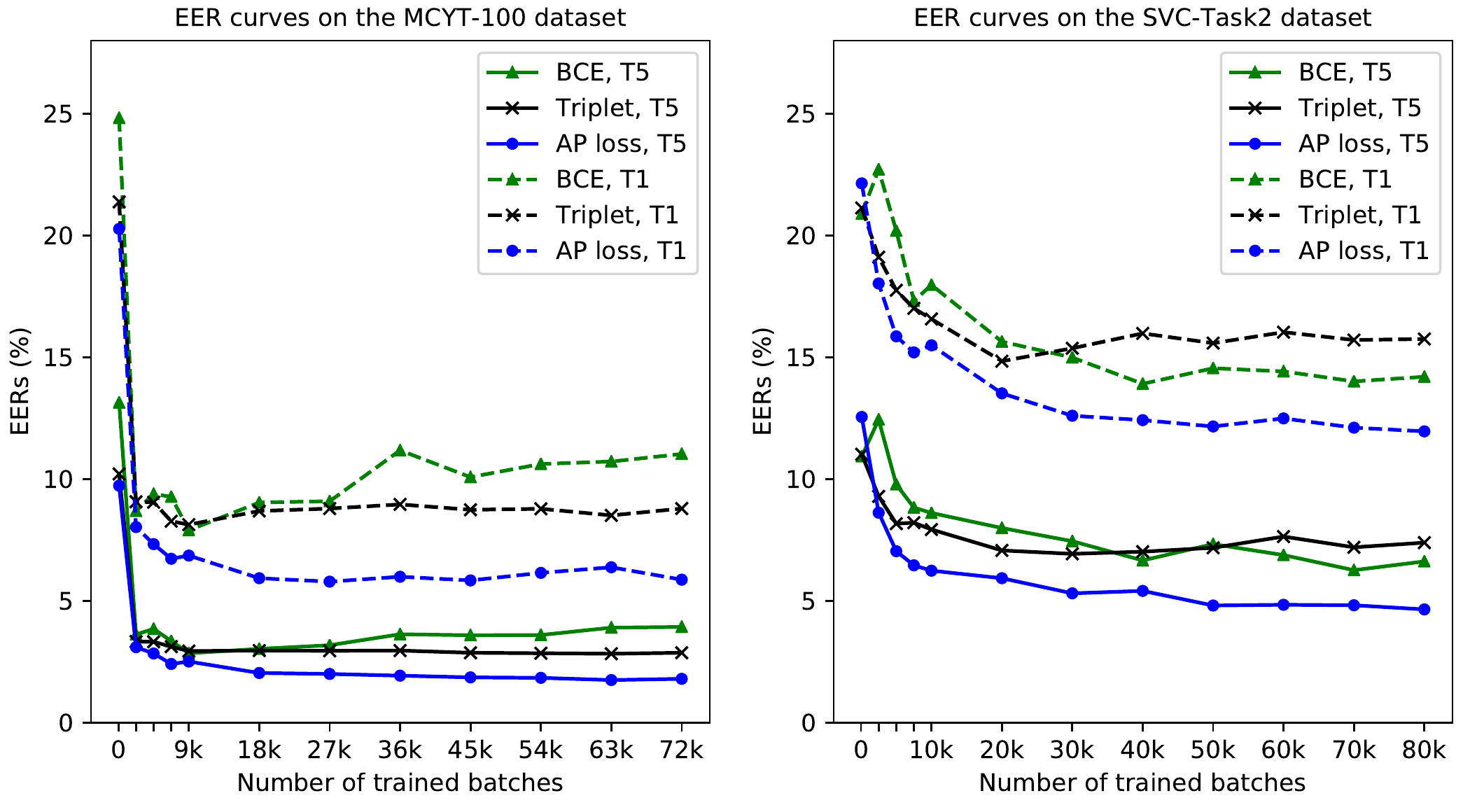}
  \caption{EER curves on the MCYT-100 and the SVC-Task2 datasets for AP loss, triplet loss and BCE.}\label{figTestCurve}
\end{figure*}

First, based on synthetic signatures, we compared AP loss with triplet loss and binary cross-entropy (BCE) to examine its properties. For triplet loss, the distance metric in Eq. \ref{distance} was used. Eight hardest triplets were mined from a total of $|\mathcal{G}_1|\times|\mathcal{G}_2|=50$ triplets, and the margin was set as 0.25; pair-wise distances within $\mathcal{G}_1$ were also added to the loss to minimize the intra-class variance. For BCE, cosine similarity in Eq. \ref{cosineSimilarity} was computed, rescaled, and activated by the sigmoid function. Within a batch, there were $|\mathcal{G}_1|=5$ positive and $|\mathcal{G}_2|=10$ negative pairs, which were labeled as (0.9, 0.1) and (0.5, 0.5) respectively. The loss of positive pairs was doubled for balance.

\begin{table}[tb]
\caption{Performance in EER (\%) for three loss functions.}\label{tableLosses}
  \centering
  \begin{tabular}{cccc}
    \hline
    \multirow{2}{*}{Losses}&\multirow{2}{*}{Datasets}&\multicolumn{2}{c}{Scenarios}\cr
    \cline{3-4}
    &&T5&T1\\
    \hline
    BCE & \multirow{3}{*}{MCYT-100} & 2.85 & 7.90 \\
    Triplet & & 2.83 & 8.51\\
    AP loss & & \textbf{1.71} & \textbf{5.50} \\
    \hline
    BCE & \multirow{3}{*}{SVC-Task2} & 6.26 & 14.01\\
    Triplet & & 6.93 & 15.37\\
    AP loss & & \textbf{4.65} & \textbf{11.96} \\
    \hline
  \end{tabular}
\end{table}

The EER curves (using a global threshold), as functions of the number of trained batches, are shown in Fig. \ref{figTestCurve}. First, we can see that the AP loss consistently outperformed the other two losses. Second, the AP loss was more robust against overfitting, and continued to decrease the EERs in late training iterations. Third, somewhat surprisingly, BCE and the triplet loss presented different behaviors on two datasets. A possible reason is that, the triplet loss involved much complexity in hard sample mining and determining a proper margin, which should be carefully treated for different datasets. Detailed EERs are given in Table \ref{tableLosses}. On the SVC-Task2 dataset, previous best EERs are 7.80\% in scenario T5 and 18.25\% in scenario T1, and our method reduces the EERs by 40.4\% and 34.5\%, respectively. On the MCYT-100 dataset, our method reduces the EERs by 5.6\% and 59.4\% in scenarios T5 and T1, respectively. The great performance improvement demonstrates that our model could extract intrinsic and robust representations from real-world signatures via learning from synthesized ones.

\begin{figure*}
  \centering
  \includegraphics[width=0.8\textwidth, height=0.35\textwidth]{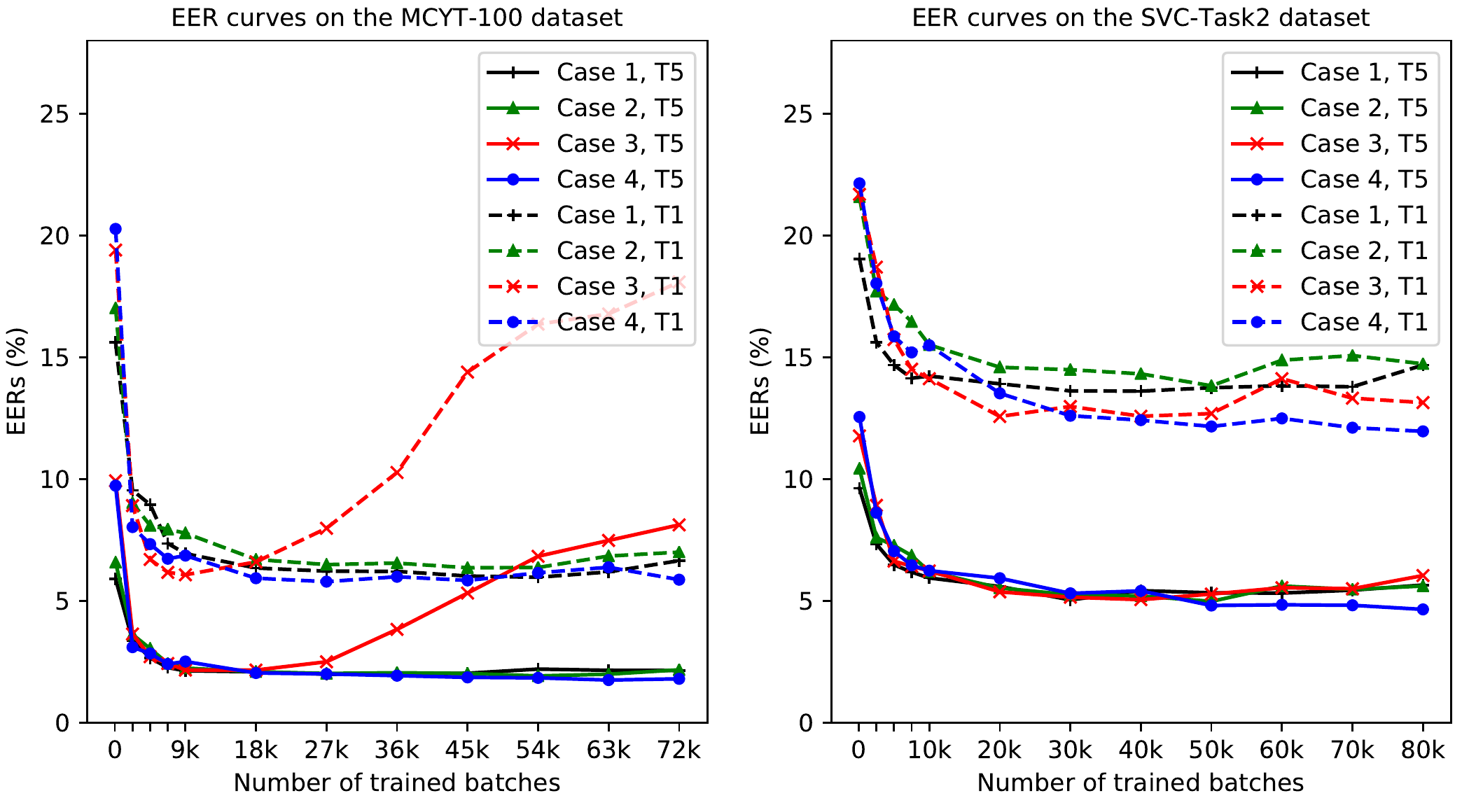}
  \caption{EER curves on the MCYT-100 and the SVC-Task2 datasets for different usage cases of synthetic signatures.}\label{figTestCurve2}
\end{figure*}

We further compared synthetic signatures with real handwritten ones. Specifically, we compared the following cases:
\begin{enumerate}
  \item Signatures in $\mathcal{G}_1$ and $\mathcal{G}_2$ were replaced with genuine signatures and skilled forgeries, respectively;
  \item Signatures in $\mathcal{G}_2$ were replaced with skilled forgeries;
  \item Signatures in $\mathcal{G}_1$ were replaced with genuine signatures;
  \item Signatures in both $\mathcal{G}_1$ and $\mathcal{G}_2$ were synthetic.
\end{enumerate}
All models were trained in exactly the same way using the AP loss, and the EER curves are shown in Fig. \ref{figTestCurve2}. There were two important observations. First, we can see that synthesized signatures were even more effective than real handwritten signatures for training, because they were constructed from disturbed $\Sigma\Lambda$ parameters and therefore tightly bounded the template signatures. Second, case 3 converged the fastest, but led to slight overfitting on SVC-Task2 and severe overfitting on MCYT-100. Therefore, when using case 3, a representative validation set is necessary; when using case 4, we can simply train for a large number of batches and use the final models, which are generally also the best-performing ones. Detailed EERs are given in Table \ref{tableSF}.

\begin{table}
\caption{Performance in EER (\%) for different usage cases of synthetic signatures.}\label{tableSF}
  \centering
  \begin{tabular}{ccccc}
    \hline
    \multicolumn{2}{c}{Synthetic?}&\multirow{2}{*}{Datasets}&\multicolumn{2}{c}{Scenarios}\cr
    \cline{1-2}\cline{4-5}
    $\mathcal{G}_1$&$\mathcal{G}_2$&&T5&T1\\
    \hline
    X & X & \multirow{4}{*}{MCYT-100} & 1.99 & 6.21\\
    $\checkmark$ & X & & 1.92 & 6.37\\
    X & $\checkmark$ & & 2.16 & 6.07 \\
    $\checkmark$ & $\checkmark$ & & \textbf{1.71} & \textbf{5.50}\\
    \hline
    X & X & \multirow{4}{*}{SVC-Task2} & 5.04 & 13.62 \\
    $\checkmark$ & X & & 4.98 & 13.83 \\
    X & $\checkmark$ & & 5.05 & 12.58 \\
    $\checkmark$ & $\checkmark$ & & \textbf{4.65} & \textbf{11.96} \\
    \hline
  \end{tabular}
\end{table}

\subsection{Comparison with state-of-the-art}
\begin{table*}[tb]
\caption{Comparison of EERs (\%) with state-of-the-art methods on the MCYT-100 and the SVC-Task2 datasets.}\label{comparison}
  \centering
  \begin{tabular}{ccccc}
    \hline
    \multirow{2}{*}{Datasets} & \multirow{2}{*}{Methods} & Number of & \multicolumn{2}{c}{Threshold}\\
    \cline{4-5}
    &&templates&Global & User-specific\\
    \hline
    \multirow{7}{*}{MCYT-100} & SRSS based on $\Sigma\Lambda$ model \cite{diaz2016dynamic} & 1 & 13.56 & - \\
    & SynSig2Vec (Ours) & 1 & \textbf{5.50} & \textbf{2.15} \\
    \cline{3-5}
    & Symbolic representation \cite{guru2017interval} & 5 & 5.70 & 2.20 \\
    & DTW warping path score \cite{sharma2017exploration} & 5 & 2.76 & 1.15 \\
    & DTW with SCC \cite{xia2017signature} & 5 & - & 2.15 \\
    & Recurrent adaptation networks \cite{lai2018recurrent} & 5 & 1.81 & - \\
    & SynSig2Vec (Ours) & 5 & \textbf{1.71} & \textbf{0.93} \\
    \hline
    \multirow{5}{*}{SVC-Task2} & SRSS based on $\Sigma\Lambda$ model \cite{diaz2016dynamic} & 1 & 18.25 & - \\
    & SynSig2Vec (Ours) & 1 & \textbf{11.96} & \textbf{7.34}\\
    \cline{3-5}
   %& SVC-competition \cite{yeung2004svc2004} & 5 & - & 5.50 \\
    & DTW warping path score \cite{sharma2017exploration} & 5 & 7.80 & \textbf{2.53} \\
    & DTW with SCC \cite{xia2017signature} & 5 & - & 2.63 \\
    & SynSig2Vec (Ours) & 5 & \textbf{4.65} & 2.63\\
    \hline
  \end{tabular}
\end{table*}

In Table \ref{comparison}, we compare our method with state-of-the-art methods on the MCYT-100 and the SVC-Task2 datasets. Our method has achieved substantial improvements over the previous methods, especially in scenario T1 where only one template is available. In scenario T1, our method reduces the EERs by 59.4\% (=(13.56-5.50)/13.56*100\%) and 34.5\% (=(18.25-11.96)/18.25*100\%) on the MCYT-100 and the SVC-Task2 datasets, respectively.

\subsection{Limitations and future work}
Some issues in SynSig2Vec need further study, such as the effects of signature distortion levels and the number of synthetic signatures. Besides, SynSig2Vec only uses the pen-down components and is expected to have further improvements in future work by considering the pen-ups if available.

\section{Conclusion}
In this paper, we propose to learn dynamic signature representations through ranking synthesized signatures. The $\Sigma\Lambda$ model is introduced to synthesize two groups of signatures for each given genuine signature. Signatures in the first group have lower distortion levels and should rank higher according to the similarity to the template signature, as compared to those in the second group. We construct a lightweight one-dimensional CNN to learn to rank these synthesized samples, and incorporate the AP of ranking into the loss function for optimization. Our method only requires genuine signatures for training, yet substantially improves the state-of-the-art performance on two public benchmarks. Particularly, when only one template signature is available for the verifier, our method surpasses the state-of-the-art on the MCYT-100 benchmark by 8.06\%, and on the SVC-Task2 benchmark by 6.29\%, showing its significant effectiveness and great potential.

\section{Acknowledgement}
This research is supported in part by NSFC (Grant No.: 61936003), the National Key Research and Development Program of China (No. 2016YFB1001405), GD-NSF (no.2017A030312006), Guangdong Intellectual Property Office Project (2018-10-1), and GZSTP (no. 201704020134).

\bibliography{refs.bib}
\bibliographystyle{aaai}

\end{document}